%% file: iclr2025_conference.tex
\documentclass{article} % For LaTeX2e

\usepackage{microtype}
\usepackage{graphicx}
\usepackage{subcaption}
\usepackage{csquotes}
\usepackage{afterpage}
\usepackage{xcolor}

\usepackage{cuted}
\usepackage{caption}

\usepackage{amsmath}
\usepackage{amssymb}
\usepackage{mathtools}
\usepackage{amsthm}
\usepackage{xspace}
\usepackage{colortbl}
\usepackage{multirow}
\usepackage{enumitem}
\usepackage{wrapfig}
\usepackage{nicefrac}
\usepackage[font=small,labelfont=bf,justification=centering]{caption}

\usepackage{booktabs}
\usepackage{multirow}
\usepackage{graphicx}
\usepackage{xcolor}
\usepackage{pifont} % For checkmarks and crosses
\usepackage{threeparttable}

\usepackage{geometry}
\usepackage{tabularx}

\usepackage{amsmath}

\usepackage[T1]{fontenc}

\usepackage[utf8]{inputenc}

\usepackage{inconsolata}

\usepackage{graphicx}

\usepackage{booktabs}
\usepackage{multirow}
\usepackage{graphicx}
\usepackage{amsmath}
\usepackage{amssymb}
\usepackage{xspace}
\usepackage{color, colortbl}

\makeatletter
\DeclareRobustCommand\onedot{\futurelet\@let@token\@onedot}
\def\@onedot{\ifx\@let@token.\else.\null\fi\xspace}

\def\eg{\emph{e.g}\onedot} 
\def\ie{\emph{i.e}\onedot}

\makeatother

\definecolor{LightPurple}{RGB}{240,236,245}
\newcommand{\RC}[1]{\rowcolor{LightPurple}}

\usepackage{enumitem}

\newcommand{\Method}{{\bfseries\scshape Apollo}\xspace}
\usepackage{iclr2025_conference,times}

\input{math_commands.tex}

\usepackage{hyperref}
\usepackage{url}

\title{Unified Multi-Task Audio-Video Joint Generation}

\author{
  Jun Wang\thanks{Equal contribution} \quad
  Chunyu Qiang\footnotemark[1] \quad
  Yuxin Guo \quad
  Yiran Wang \quad
  Xijuan Zeng \quad
  Feng Deng\\
  Kuaishou Technology \\
  \texttt{\{wangjun06, qiangchunyu\}@kuaishou.com}
  }

\iclrfinalcopy % Uncomment for camera-ready version.
\begin{document}

\renewcommand{\thefootnote}{\fnsymbol{footnote}}

\maketitle

% ==================================================================
% Teaser 图片插入位置：直接位于 \maketitle 之后，\begin{abstract} 之前
% ==================================================================
\begin{center}
    % 1. 插入图片
    \includegraphics[width=1.0\textwidth,keepaspectratio]{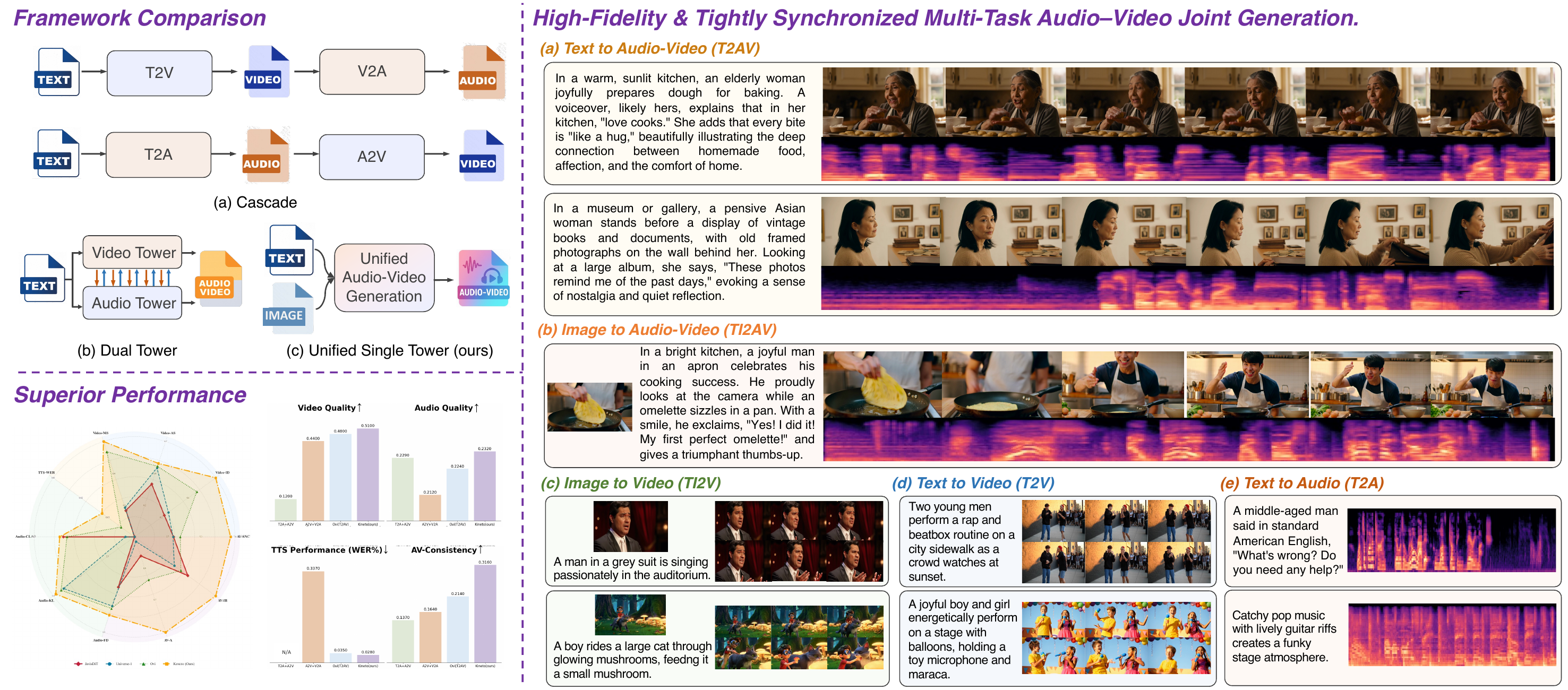}
    
    % 2. 可选：添加图片标题（不需要可删除以下两行）
    \vskip 1mm % 图片与标题的间距
    \captionof{figure}{
    We propose \Method, a unified audio–video generation framework which delivers high fidelity, strong semantic and temporal alignment, and reliable instruction following in both joint and unimodal settings, with robust OOD generalization. Across tasks (T2AV/TI2AV/TI2V/T2V/T2A), it attains performance comparable to Veo-3 among open-source models.
    \label{fig:teaser} % 添加 Label 以便引用
    }
    
    % 4. 与下方 Abstract 的间距
    % \vskip 5mm
\end{center}
% ==================================================================

\begin{abstract}
Audio–video joint generation has progressed rapidly, yet substantial challenges still remain. Non-commercial approaches still suffer audio-visual asynchrony, poor lip–speech alignment, and unimodal degradation, which can be stemmed from weak audio–visual correspondence modeling, limited generalization, and scarce high-quality dense-caption data.
To address these issues, we introduce \Method and delve into three axes—model architecture, training strategy, and data curation. 
Architecturally, we adopt a single-tower design with unified DiT blocks and an Omni-Full Attention mechanism, achieving tight audio–visual alignment and strong scalability.
Training-wise, we adopt a progressive multitask regime—random modality masking to joint optimization across tasks, and a multistage curriculum, yielding robust representations, strengthening A-V aligned world knowledge, and preventing unimodal collapse.
For datasets, we present the first large-scale audio–video dataset with dense captions, and introduce a novel automated data-construction pipeline which annotates and filters millions of diverse, high-quality, strictly aligned audio–video–caption triplets.
Building on this, \Method scales to large datasets, delivering high-fidelity, semantically and temporally aligned, instruction-following generation in both joint and unimodal settings while generalizing robustly to out-of-distribution scenarios. Across tasks, it substantially outperforms prior methods by a large margin and achieves performance comparable to Veo 3, offering a unified, scalable path toward next-generation audio–video synthesis.
\end{abstract}

\input{sec/1_intro}
\input{sec/2_relatedworks}
\input{sec/3_methods}
\input{sec/4_datasets}
\input{sec/5_experiments}

\input{sec/6_conclusion}

% \section*{Acknowledgement}
% We acknowledge the contributions from (sorted by first name): Boyuan Jiang, Chen Zhang, Feng Deng, Jiahao Wang, Jingbin He, Jingke Li, Jingru Zhao, Junjie Yan, Liang Hou, Lingyu Zou, Ming Wen, Nan Li, Peihan Li, Pengfei Wan, Teng Ma, Xiaoyu Shi, Xijuan Zeng, Xin Tao, Xu Li, Yan Zhou, Yiran Wang, Yu Zhao, Yuan Gao, Yun Li, Yushen Chen, Yuxin Guo, Yuzhe Liang, Zewen Song, Zhongliang Liu, Zihan Li, Zihao Ji, Ziyang Yuan, and Ziyu Zhang.

\bibliography{iclr2025_conference}
\bibliographystyle{iclr2025_conference}

\end{document}

%% file: math_commands.tex
\usepackage{amsmath,amsfonts,bm}

\def\eqref#1{equation~\ref{#1}}
\def\1{\bm{1}}

\DeclareMathAlphabet{\mathsfit}{\encodingdefault}{\sfdefault}{m}{sl}
\SetMathAlphabet{\mathsfit}{bold}{\encodingdefault}{\sfdefault}{bx}{n}

%% file: sec/1_intro.tex
\section{Introduction}
\label{sec:intro}

In ancient Greek mythology, \emph{Apollo} presides over music, poetry, and prophecy, embodying artistic creation; he was later imbued with solar iconography and depicted as driving the sun chariot to illuminate the world. While for human perception and narrative, “illumination” arises not only from visible light and shadow but also from audible sensation: vision furnishes structure and space, while hearing shapes rhythm and affect, and their interplay yields a fully immersive experience.
Accordingly, vision and auditory are complementary and jointly indispensable modalities for depicting the real world. Consequently, audio–video joint generation has emerged as one of the most prominent trends in generative AI. Commercial systems such as Veo 3 and Sora 2 have achieved impressive performance with strong semantic alignment, while diverse open-source models with varying architectures are also emerging. Nevertheless, research on joint audio–visual generation remains nascent, and current models—even some commercial systems—still exhibit audio–visual asynchrony, lip–speech mismatches, and degradation in unimodal quality. We attribute several factors: (1) Architecture. Most T2AV models like JavisDiT~\citep{liu2025javisdit}, UniVerse-1~\citep{wang2025universe} and Ovi~\citep{low2025ovi} employ a single-tower architecture with a cross-attention module, resulting in limited audio-video interaction and alignment. (2) Training strategies and data. Currently, mainstream methods perform single-task training, which might bring about biased representations and struggle to exploit underlying audio-video correlation and world knowledge.

To this end, informed by the above observations, we identify several key contributing factors.
(1) Architecturally, prevailing designs hinder thorough cross-modal interaction: most existing models adopt dual-tower architectures with modality-specific initialization, learn each modality independently, and rely on shallow fusion via cross-attention or adapters, which fails to fully align audio–visual features.
(2) For data construction, there is a pronounced lack of diverse, high-quality, and densely annotated audio–video aligned generation datasets, as well as scalable, high-quality annotation methodologies for constructing them.
(3) From a learning-strategy perspective, most existing methods are trained exclusively on text-to-AV generation. This single-task regime induces overfitting and representation bias, which in turn hinders generalization and degrades unimodal performance.

To address these issues, we propose \Method, which introduces coordinated improvements at the architectural, learning-strategy, and data levels. 
Specifically, on the architectural side, we adopt a single-tower backbone with unified DiT blocks. Each block integrates an Omni-Full Attention module that jointly attends to four streams—audio, audio captions, video, and video captions—thereby facilitating cross-modal fusion and interaction, achieving tight A-V alignment and stronger coupling to textual conditions, which offers a higher scaling ceiling.
For training, we design a progressive multitask training regime: random modality masking sustains joint optimization over T2AV/TI2AV/TI2V/T2V/T2A. A performance-adaptive pretrain–post train curriculum tunes data mixtures and quality, yielding robust, generalizable representations that exploit A/V correlations and world knowledge.
From the data construction, we introduce an automated annotation pipeline that enables efficient model scaling.
Building on these three insights, our model achieves strong performance in both joint and unimodal generation, delivering high fidelity, tight audio–visual alignment, natural outputs, and robust instruction following with favorable scaling behavior. On the Verse-Bench, it surpasses prior methods by a large margin and generalizes well to out-of-distribution (OOD) scenarios. Our main contributions are summarized as follows:
\begin{itemize}[left=2pt]
    \vspace{-3pt}
    \item{We introduce \Method, a unified framework for multi-task audio–video joint generation that, to our knowledge, is the first model to achieve performance comparable to Veo 3. \Method effectively resolves semantic and temporal audio–visual misalignment while delivering high-fidelity generation.}
    \vspace{-3pt}
    \item{Our key technical novelties lie in the unified single-tower architecture with the omni-full attention mechanism for seamless audio–visual fusion, and a progressive multi-task training strategy that promotes generalizable representations and prevents unimodal performance degradation.}
    \vspace{-3pt}
    \item We propose a large-scale, high-quality audio-video dataset consisting of 81 million samples with accurate dense captions, along with an automated data generation pipeline and a high-quality audio-video generation dataset.
    \vspace{-3pt}
    \item{Extensive experiments demonstrate that \Method consistently excels in both unimodal and joint audio–video generation, outperforming prior state-of-the-art on the unimodal benchmark and the AV joint-generation benchmark consistently.}
    
\end{itemize}

%% file: sec/2_relatedworks.tex
\section{Related Works}
\label{sec:relatedworks}

\textbf{Text-to-Video Generation (T2V).}
% Diffusion models catalyzed a step change in video generation, such as AnimateDiff~\citep{guo2023animatediff} and Video Diffusion Models~\citep{ho2022video}.
% Subsequent Stable Video Diffusion~\citep{blattmann2023stable} highlighted that large, high-quality curated datasets are critical for performance. Early approaches commonly relied on U-Net backbones. A major shift arrived with Sora~\citep{liu2024sora}, which popularized the Diffusion Transformer (DiT) architecture~\citep{peebles2023scalable} trained on massive, high-quality video corpora. CogVideoX~\citep{yang2024cogvideox}, HunyuanVideo~\citep{kong2024hunyuanvideo}, the WAN series~\citep{wan2025wan}, and Step-Video~\citep{ma2025step} released several open-source models, alongside strong closed-source systems such as the Kling series~\citep{wang2025kling}, SeeDance 1.0~\citep{gao2025seedance}, MovieGen~\citep{polyak2024movie}, and Veo 2~\citep{deepanway2023text}. Architecturally, most contemporary models share a common blueprint: a 3D VAE compresses videos into a spatiotemporal latent space, and a DiT performs denoising in that space. Across these systems, data quality and scale remain paramount, rendering data curation and processing central to model development.
Diffusion models have revolutionized video generation, with AnimateDiff~\citep{guo2023animatediff} and Video Diffusion Models~\citep{ho2022video} leading the way. Stable Video Diffusion~\citep{blattmann2023stable} emphasized the importance of large, high-quality datasets for performance. Early models used U-Net backbones, but Sora~\citep{liu2024sora} introduced the Diffusion Transformer (DiT) architecture~\citep{peebles2023scalable}, trained on extensive video corpora. Open-source models like CogVideoX~\citep{yang2024cogvideox}, HunyuanVideo~\citep{kong2024hunyuanvideo}, and the WAN series~\citep{wan2025wan} joined closed-source systems like Kling~\citep{wang2025kling} and Veo 2~\citep{deepanway2023text}. Most models share a common architecture: a 3D VAE compresses videos into spatiotemporal latents, and a DiT performs denoising. Data quality and scale remain crucial for model success, highlighting the importance of data curation and processing.

% Diffusion models have revolutionized video generation, with AnimateDiff~\citep{guo2023animatediff} and Video Diffusion Models~\citep{ho2022video} leading the way. Stable Video Diffusion~\citep{blattmann2023stable} highlighted the need for large, high-quality datasets. Early models used U-Net backbones, but Sora~\citep{liu2024sora} introduced the Diffusion Transformer (DiT) architecture~\citep{peebles2023scalable} trained on vast video corpora. Open-source models like CogVideoX~\citep{yang2024cogvideox}, HunyuanVideo~\citep{kong2024hunyuanvideo}, and WAN~\citep{wan2025wan} joined closed-source systems like Kling~\citep{wang2025kling} and Veo 2~\citep{deepanway2023text}. Most models use a common architecture: a 3D VAE compresses videos into spatiotemporal latents, with a DiT performing denoising. Data quality and scale are crucial for success, emphasizing the importance of data curation.

\textbf{Image-to-Video Generation (I2V).}
% Early I2V systems extended T2V by conditioning on a single input frame through latent concatenation or CLIP-based feature injection~\citep{liu2023revisiting}. Subsequent work introduced more structured pipelines. Cascaded designs like I2VGen-XL~\citep{zhang2023i2vgen} employ a low-resolution semantic-preserving base stage followed by a high-resolution refinement stage. Dual-injection frameworks like DynamiCrafter~\citep{xing2024dynamicrafter} merge CLIP-projected context with direct image-noise concatenation to jointly capture motion plausibility and appearance fidelity.
% %A parallel direction focuses on lightweight integration.
% Adapter-based approaches such as LAMP~\citep{notomi2015loop} and I2V-Adapter~\citep{guo2024i2v} insert cross-frame attention modules into pre-trained T2V backbones without modifying weights, efficiently propagating first-frame features while reducing costs. More recent efforts emphasize explicit motion modeling and stable sampling like Motion-I2V~\citep{shi2024motion} and FrameBridge~\citep{wang2024framebridge}. In essence, I2V research is shifting from early single-frame conditioning toward more structured, motion-aware, and controllable frameworks, yet remains fundamentally constrained by its reliance on large curated datasets, limited long-range motion modeling, and persistent tension between appearance fidelity and motion realism.
Early I2V systems extended T2V by conditioning on a single frame via latent concatenation or CLIP-based feature injection~\citep{liu2023revisiting}. Later, cascaded designs like I2VGen-XL~\citep{zhang2023i2vgen} and dual-injection frameworks like DynamiCrafter~\citep{xing2024dynamicrafter} introduced structured pipelines for better motion and appearance fidelity. Adapter-based approaches like LAMP~\citep{notomi2015loop} and I2V-Adapter~\citep{guo2024i2v} integrate cross-frame attention. Recent works on motion modeling and stable sampling include Motion-I2V~\citep{shi2024motion} and FrameBridge~\citep{wang2024framebridge}. Despite these advancements, I2V still faces challenges with curated datasets, limited long-range motion modeling, and the trade-off between appearance fidelity and motion realism.

\textbf{Text-to-Audio Generation (TTA).}
% Recent advances in generative models have significantly advanced text-to-audio generation. Make-An-Audio~\citep{huang2023make} and AudioLDM~\citep{liu2023audioldm,liu2024audioldm}, synthesize audio through iterative denoising of text-conditioned latent representations. Tango~\citep{liu2024tango, hung2024tangoflux}, Audio Flamingo~\citep{kong2024audio}, GenAu~\citep{haji2024taming}, Fugatto~\citep{valle2025fugatto} further enhance design spaces of latent space, data quality and cross-modal alignments. Recently, Stable Audio series~\citep{evans2025stable} employs hierarchical latent diffusion trained on large-scale datasets for high-fidelity output. Beyond diffusion-based priors, flow-matching techniques optimize probability density transport for audio synthesis. VoiceBox~\citep{le2023voicebox} enables zero-shot style transfer via continuous normalizing flows. 
% TangoFlux~\citep{hung2024tangoflux} introduces CLAP-ranked preference optimization to enhance text-audio alignment. 
% Existing methods align text and audio semantically but primarily target descriptive queries, limiting interactive control and adaptability to evolving instructions.
% They are also confined to short audio domains. These limitations demand TTA models to handle complex instructions over long durations.
Recent advances in generative models have propelled text-to-audio generation. Models like Make-An-Audio~\citep{huang2023make} and AudioLDM~\citep{liu2023audioldm} synthesize audio via iterative denoising of text-conditioned latent representations. Tango~\citep{liu2024tango}, Audio Flamingo~\citep{kong2024audio}, and others expand latent spaces and enhance cross-modal alignment. Stable Audio~\citep{evans2025stable} employs hierarchical latent diffusion for high-fidelity output. AudioStory~\citep{guo2025audiostory} introduces a unified generation framework, achieving long-form audio generation for the first time, while flow-matching techniques like VoiceBox~\citep{le2023voicebox} enable zero-shot style transfer. TangoFlux~\citep{hung2024tangoflux} optimizes text-audio alignment with CLAP-ranked preferences. Though these methods excel at semantic alignment, they are limited to short durations and lack flexibility for complex, evolving instructions, underscoring the need for TTA models that handle long, complex tasks.

\textbf{Audio-Video Joint Generation (T2AV).}
Pioneering efforts like MM-Diffusion~\citep{ruan2023mm} use coupled U-Net backbones, while DiT-based approaches dominate. AV-DiT~\citep{wang2024av} adapts pre-trained image DiTs with lightweight adapters, and UniForm~\citep{zhao2025uniform} uses a unified single-tower architecture for audio-video tokens. Key challenges include precise spatio-temporal synchronization, addressed by methods like JavisDiT~\citep{liu2025javisdit} (hierarchical prior), Ovi~\citep{low2025ovi} (twin-backbone design), and SyncFlow~\citep{liu2024syncflow} (dual-DiT with Rectified Flow Matching). Other research orchestrates unimodal experts, such as MMDisCo~\citep{hayakawa2024mmdisco} and Universe-1~\citep{wang2025universe}, which combine specialized models at the block level. Despite advances in architecture and data, most models focus on sound effects or music, leaving synchronized speech and video synthesis an underexplored challenge.

%% file: sec/3_methods.tex
\section{\Method}
\label{sec:methods}

\begin{figure*}[!t]
    \centering
    \includegraphics[width=1.0\linewidth]{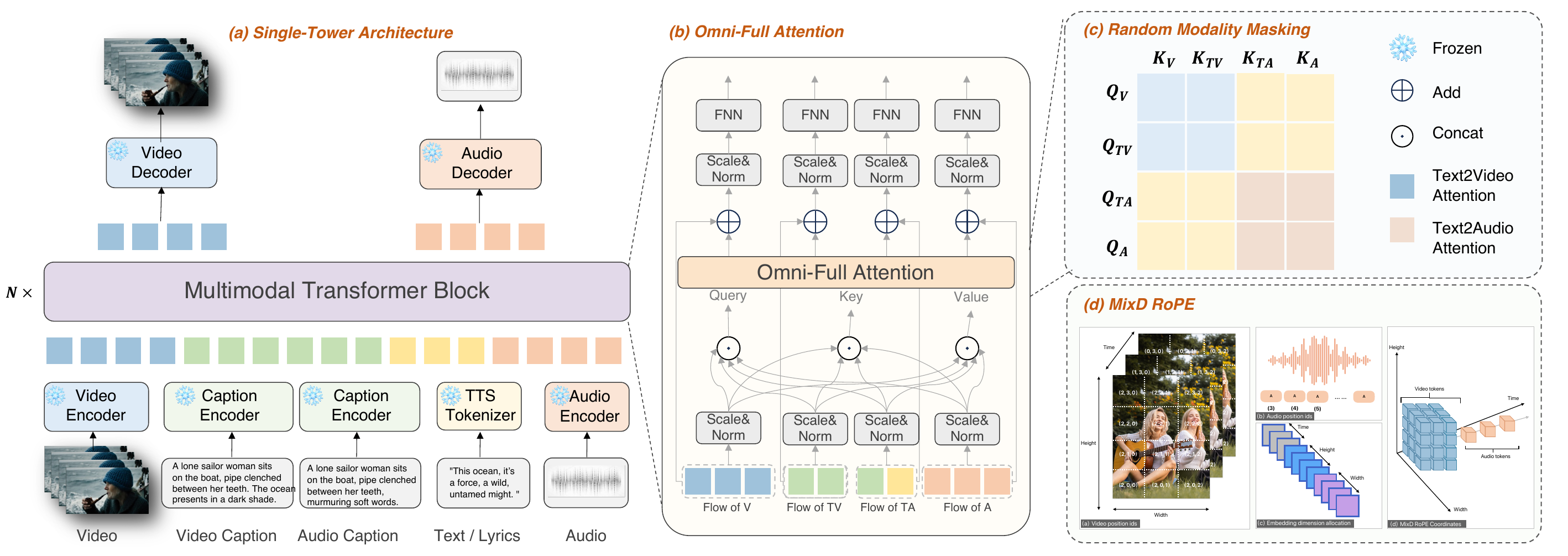}
    \vspace{-10pt}
    \caption{Overview of \Method. The model takes four inputs: video, video-related text, audio-related text, and audio. Each input is individually encoded by respective encoders, then fed into the MM-DiT. The MM-DiT module outputs the latent variables of video and audio, which are then decoded separately into video and audio.}
    \vspace{-10pt}
    \label{Overall}
\end{figure*}

\subsection{Preliminary}

\textbf{Problem Definition.}
Our goal is to enable the generation of both audio and video within a single model, given various prior conditions. We denote the denoising network as $\epsilon_{\theta}(\cdot)$, the text condition as $c$. Let $\{z^a_t\}_{t\in[0,1]}$ and $\{z^v_t\}_{t\in[0,1]}$ denote the latent variables at timestamp $t$ for audio and video, respectively. Here, $t=0$ denotes the final timestamp of pure Gaussian noise. During inference, $\epsilon_{\theta}(\cdot)$ recursively performs denoising from $t=0$ to $t=1$ to produce the final generation, $\hat{z}^a_1, \hat{z}^v_1$, as shown below:
\begin{equation}
    \hat{z}^a_{t^\prime}, \hat{z}^v_{t^\prime} = \epsilon_\theta(z^a_t,z^v_t,t,c),
\end{equation}

\textbf{Conditional Flow-Matching.}
We employ flow matching as the denoising objective. The model needs to learn the velocity field that transforms pure noise $p_0=\mathcal{N}(0,\boldsymbol{I})$ to the underlying data distribution $p_\text{data}$. In practice, we perform linear interpolation $x_t=(1-t)x_0+tx_1$ to construct a distribution at timestamp $t$. Here, $x_0\sim p_0$ and $x_1\sim p_\text{data}$. Given the condition $c$, the model $\epsilon_{\theta}(\cdot)$ is trained to predict the target velocity, \ie, constantly as $u_t=x_1-x_0$:
\begin{equation}
\begin{aligned}
    \mathcal{L}_\text{FM}&=\mathbb{E}_{t,c,{x}_0,{x}_1}\Big\Vert ({x}_1-{x}_0)-\epsilon_\theta\big(t{x}_1+(1-t){x}_0,t,c\big) \Big\Vert_2^2, \\
    & \text{where}\quad t\sim\mathcal{U}(0,1),\ {x}_0\sim \mathcal{N}(0,\boldsymbol{I}),\ {x}_1\sim p_\text{data}.
    \label{eq:flow}
\end{aligned}
\end{equation}

\textbf{Latent Encoding.}
The model takes four inputs: video, video-related text, audio-related text, and audio, where video-related text represents the video caption and audio-related text represents the audio caption and speech text. Video is encoded by the 3d casual visual encoder from CogVideoX~\citep{yang2024cogvideox},  
We use Qwen3-8B Embedding~\citep{zhang2025qwen3} as the encoder for audio and video captions.

\subsection{Single Tower with Full Attention}

\textbf{Single Tower DiT.}
To ensure a thorough audio-video fusion, we employ a single-tower architecture. As shown in Fig.~\ref{Overall}, following Stable Diffusion 3~\citep{esser2024scaling}, we employ Multimodal Diffusion (MMDiT) to take the sequences of all modalities as input and perform full attention. Specifically, there are four inputs, \ie, video, video-related text, audio-related text, and audio. Each type of input is individually encoded into latents with respective encoders, then fed into the MM-DiT. The MM-DiT module outputs the latent variables of video and audio in two streams, which are then decoded separately to perform video and audio generation.

% \vspace{-3pt}
\textbf{Mixed Dimension Rotary Position Embedding (MixD-RoPE).}
Another key architectural innovation is Mixed Dimension Rotary Position Embedding (MixD-RoPE).
% Unlike the traditional 1D-RoPE , which is limited to encoding one-dimensional positional information, M-RoPE effectively models the positional information of multimodal inputs. This is achieved by deconstructing the original rotary embedding into three components: temporal, height, and width. For audio inputs, these components utilize identical position IDs, making M-RoPE functionally equivalent to 1D-RoPE (Su, 2024). When processing videos, the temporal IDs of each visual frame remain constant, while distinct IDs are assigned to the height and width components based on the position in certain frame. For video, which is treated as sequences of frames, the temporal ID increments for each frame, while the height and width components follow the same ID assignment pattern as images. 
As shown in Fig.~\ref{Overall} (d), to enhance the positional information introduced by various aspect ratios and duration in videos, we apply 3D RoPE encoding across three dimensions, \ie, temporal, width and height for video embedding. This 3D RoPE incorporates both absolute and relative position dependency in videos. For audio modality, we employ compatible temporal 1D positional encodings, while its position number is initialized by incrementing the maximum temporal position ID of the video modality by one. As a result, we build MixD-ROPE with a shared temporal position ID between video and audio modalities.

%To further facilitate the effective fusion of positional information between audio and videos, inspired by Qwen2-VL \cite{}, we build MixD-ROPE in the concatenated sequences by adding compatible 1D positional encoding for audio embeddings, where three components share the same position ID. Specifically, the position numbering for audio embeddings is initialized by incrementing the maximum time position ID of video embedding by one.
% \begin{figure}
%     \centering
%     \includegraphics[width=0.95\linewidth]{RoPEV1.pdf}
%     \caption{Illustration of M-RoPE. (a) The video position ids, including temporal, width and height components. (b) The audio position ids, where three components share the same position ID and are initialized by incrementing the maximum time position ID of video embedding by one. (c) Embedding dimension allocation of temporal, width and height components. (d) 3D RoPE coordinates. Audio embedding follows video embedding along the space diagonal.}
%     \label{RoPE}
% \end{figure}

% In scenarios where the model’s input encompasses multiple modalities, position numbering for each modality is initialized by incrementing the maximum position ID of the preceding modality by one.

% \vspace{-3pt}
\textbf{Omni-Full Attention.}
% or \subsection{MMDiT Blocks}
Previous works may employ separated spatial and temporal attention to reduce computational complexity, like UniForm~\citep{zhao2025uniform}. However, as in CogVideoX~\citep{yang2024cogvideox}, this separate attention mechanism requires extensive implicit information transmission, significantly increasing the learning complexity.
Other works tailor two transformer towers for audio and video generation separately, \eg, AV-DiT, SyncFlow, JavisDiT, TAVGBench.% However AV-DiT and SyncFlow only used video modality to affect audio generation but failed to do so reversely.
However, they often adopt a multi-stage training approach, which is complex and resource-intensive. The two towers must first be pretrained separately, then finetuned together, increasing training time and resource consumption.
To achieve more efficient training and more effective modality fusion, we employ the 3D text-video-audio hybrid full attention mechanism.
As shown in Fig.~\ref{Overall}, within the MM-DiT module, the hidden states of video, video-related text, audio-related text, and audio are first scaled and normalized, then concatenated together for the attention calculation. 
\begin{align}
    & Q = Q_V \odot Q_{VT} \odot Q_{AT} \odot Q_{A}, \\
    & K = K_V \odot K_{VT} \odot K_{AT} \odot K_{A}, \\
    & V = V_V \odot V_{VT} \odot V_{AT} \odot V_{A}, \\
    & \text{Attn}(Q, K, V) = \text{Softmax}(\frac{QK^\top}{\sqrt{d_{k}}})V,
\end{align}
The attention values are then split into separate hidden states, which undergo scaling and normalization, residual connection, and feedforward, and subsequently fed to the next MM-DiT module. As a result, we achieve the unification of all input modalities within joint full-attention.

% \begin{figure}
%     \centering
%     \includegraphics[width=0.95\linewidth]{Figure1V3.pdf}
%     \caption{Within the MM-DiT module, the hidden states of video, video-related text, audio-related text, and audio are first scaled and normalized, then concatenated together for the attention calculation. The result is then split into separated hidden states, which undergo scaling and normalization, residual connection, and feedforward. Finally, the results are passed to the next MM-DiT module.}
%     \label{FullAttn}
% \end{figure}

\subsection{Multi-Task Progressive Training Strategy}

\textbf{Random Modality Masking.}
To learn generalizable and robust audio-visual representations for joint generation, we train the generative model with a broad spectrum of tasks.
As a result, we propose to selectively adjust the mask of query and key for audio and video modalities. If we restrict the query and key to video embedding and video caption embedding, the model degenerates to a T2V model. Similarly, limiting the query and key to audio embedding and audio text embedding results in a T2A model. In this way, the model could not only handle joint generation, but also maintain the abilities of single-modality generation. Considering the scarcity of high-quality audio-video paired data, our method offers an alternative for training the T2VA model. We first pre-train \Method on T2V and T2A tasks, and then finetune our model on audio-video paired data to finally construct a T2VA model.
The learning objectives for audio and video generation are in Eq.~\eqref{eq:loss_audio} and Eq.~\eqref{eq:loss_video}:
\begin{align}
    & \mathcal{L}_{T2A} = ||\epsilon_{t}^a - \text{Mask}_{a}(\epsilon_{\theta}(z^a_t, z^v_t,c)) ||^{2}_{2}, \label{eq:loss_audio} \\
    & \mathcal{L}_{T2V} = ||\epsilon_{t}^v - \text{Mask}_{v}(\epsilon_{\theta}(z^a_t, z^v_t,c)) ||^{2}_{2},
    \label{eq:loss_video}
\end{align}
where $\text{Mask}_a$ is used to extract the audio token from the combined noise representation and $\text{Mask}_v$ is used to extract the vision tokens.
In summary, $\mathcal{L}_{T2A}$ and $\mathcal{L}_{T2V}$ denote the single-modality tasks of T2A and T2V. To learn the generalizable and robust world knowledge of audio-visual correlation, we also incorporate several tasks of T2AV, I2V and I2AV.
Consequently, the overall multi-task learning objective is as follows:
\begin{equation}
    \mathcal{L}_{overall} = \mathcal{L}_{T2A} + \mathcal{L}_{T2V} + \mathcal{L}_{T2AV} + \mathcal{L}_{I2V} + \mathcal{L}_{I2AV}
\end{equation}

% \begin{figure}
%     \centering
%     \includegraphics[width=0.95\linewidth]{JointAttnMaskV2.pdf}
%     \caption{Illustration of joint attention mask. The left denotes the MMDiT block, the right denotes the Query and Key while computing joint attention.}
%     \label{JointAttnMask}
% \end{figure}

% \subsection{Expert Adaptive Layernorm}
% % Then how about text?
% We concatenate the embeddings of audio and video at the input stage for better fusion of visual and acoustic information. However, the feature spaces of these two modalities differ significantly, and their embeddings may even have different numerical scales. To better process them within the same sequence, we employ the Expert Adaptive Layernorm to handle each modality independently. As shown in Figure \ref{FullAttn},
% following DiT, we use the timestep t of the diffusion process as the input to the modulation module. Then, the Vision Expert Adaptive Layernorm (Vison Expert AdaLN) and Audio Expert Adaptive Layernorm (Audio Expert AdaLN) apply this modulation to the vision hidden states and audio hidden states, respectively. This strategy promotes the alignment of feature spaces across two modalities while minimizing additional parameters.

\textbf{Progressive Training Strategy.}
To efficiently train AV joint generation, we adopt a progressive multi-task learning framework with random modality masking applied throughout all stages:

\textbf{Stage-I: Pre-training.}
We pretrain the model on the large-scale, multi-scene data corpus to acquire atomic generation capabilities across all tasks, including cross-modal semantic alignment, temporal synchronization, high-fidelity audio synthesis, and precise visual feature construction, which ensures basic abilities of both single modality generation and joint generation, and provides a solid foundation for subsequent post-training.

\textbf{Stage-II: Specialized Post-training.}
We then specialize the model on its weaker abilities and tasks. Guided by evaluation metrics, we adaptively rebalance data distributions across scenarios and tasks to strengthen underperforming capabilities while preserving overall competency.

\textbf{Stage-III: Quality-Refined Post-training.}
Finally, we fine-tune the model on the manually-curated, high-quality dataset to refine generation fidelity and enhance robustness in complex scenes, yielding improvements in perceptual realism and overall generation quality.

%% file: sec/4_datasets.tex
\section{Dataset Construction}
\label{sec:datasets}

\begin{figure*}[t]
  \centering
  % \fbox{\rule{0pt}{2in} \rule{0.9\linewidth}{0pt}}
    \includegraphics[width=0.95\linewidth]{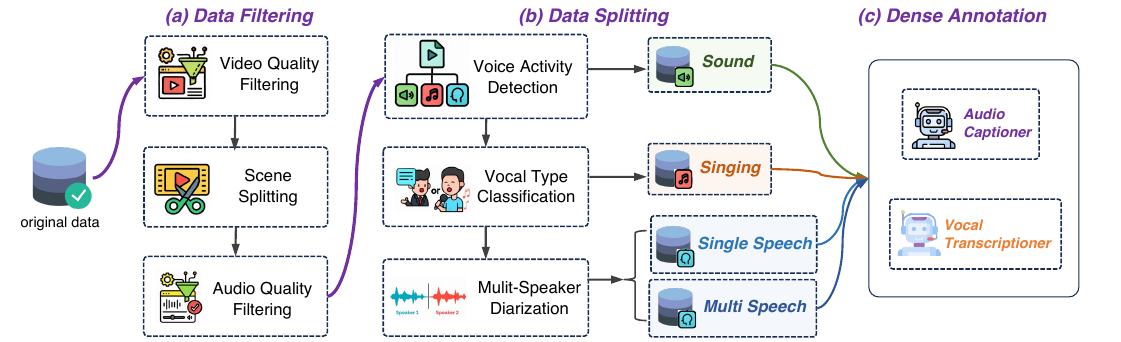}
    \vspace{-7pt}
   \caption{Overview of our Dataset Annotation Pipeline.}
   \label{fig:dataset}
   \vspace{-10pt}
\end{figure*}

% \vspace{-3pt}
\textbf{Dataset Overview.}
Our dataset comprises automatically annotated samples. The dataset contains single-speaker speech, multi-speaker speech, singing, and natural sound clips, with an overall post-filtering retention rate of 27\%.

\subsection{Dataset Filtering}
\textbf{Video Filtering and Scene Splitting.}
We first filter video quality by modeling dynamic quality (subject motion ratio, camera stability), static quality (sharpness, aesthetics, color saturation), content naturalness (no excessive effects/watermarks), and safety. We discard those videos with low resolution, low SNR/MOS, or over 20\% silence. We then apply scene splitting to ensure each sample contains only one scene.

\textbf{Audio Filtering and Post Processing.}
We filter audio data by removing samples with low SNR, MOS, abnormal clipping, distortion, or noise, ensuring less than 20\% silence, high fidelity, and consistent formatting. We then assess audio–visual consistency, using Synchformer for temporal alignment and ImageBind for semantic alignment, ensuring high synchronization in both temporal and semantic dimensions.

\subsection{Audio-Guided Data Splitting}
% We partition the dataset by audio type. We first separate vocal from non-vocal clips, forming a sound split. From the vocal subset, we further derive a singing split, a single-speaker speech split, and a multi-speaker speech split, and then apply dense captioning to each subset.

We partition the dataset by audio type, separating vocal from non-vocal clips to form a sound split. From the vocal subset, we create singing, single-speaker speech, and multi-speaker speech splits, then apply dense captioning to each.

\subsection{Dense Annotation and Integration}
% We annotate each split with specialized models to obtain speech transcripts, audio captions, and video captions, covering both meta information and detailed content. For speech and singing splits, we extract speaker attributes (e.g., gender, age, accent, style, pitch) as meta tags, while the sound split receives only audio captions. Concretely, we use Whisper-Large-v3, SenseVoice, and Qwen2.5-Omni for transcription, Qwen2.5-Omni and Gemini 2.5-Pro for audio captions, and a video expert model for fine-grained video labels. All annotations are merged into unified dense captions.

We annotate each split with specialized models for speech transcripts, audio captions, and video captions, including both meta information and detailed content. For speech and singing, we extract speaker attributes (\eg, gender, age), while the sound split receives only audio captions. We use Whisper-Large-v3, SenseVoice, and Qwen2.5-Omni for transcription, Qwen2.5-Omni and Gemini 2.5-Pro for audio captions, and a video expert model for detailed video labels. All annotations are merged into unified dense captions.

% \subsection{KinetoDense-81M for Training}
% Following the above pipeline, we obtain the 81M-sample KinetoDense dataset for pretraining and the post-train-special stage, where task-wise data ratios are adaptively adjusted based on performance. Moreover, to further improve quality, we manually annotate an additional dense-caption set with the same schema and 96\% labeling accuracy, which is used in the post-train-quality stage.

% \subsection{KinetoDense-Bench for Evaluation}
% From the above subsets, we further select a portion of samples with high audio–visual consistency for meticulous manual annotation, yielding a 320-sample test set used to evaluate semantic alignment, temporal sync, audio quality, and scene suitability. Details are provided in the appendix.

%% file: sec/5_experiments.tex
\section{Experiments}
\label{sec:experiments}

In this section, we first present the experimental setup and implementation details (Sec.~\ref{sec:exp_setup}), then compare \Method with diverse baselines across multiple tasks (Sec.~\ref{sec:compare_sota}), complemented by qualitative results (Sec.~\ref{sec:visualize}). We further conduct ablations on the unified single-tower architecture, multi-task versus progressive training, and the role of 3D RoPE for native FPS (Sec.~\ref{sec:ablation}), collectively validating the effectiveness of our approach.
% In this section, we first present the experimental setup (Sec.~\ref{sec:exp_setup}) and compare \Method with baselines(Sec.~\ref{sec:compare_sota}), supported by qualitative results (Sec.~\ref{sec:visualize}). Ablations on the single-tower architecture, multi-task and progressive training (Sec.~\ref{sec:ablation}) validate the effectiveness of our approach.

\subsection{Experimental Setup}
\label{sec:exp_setup}

\textbf{Implementation details.}
\Method comprises 26B parameters with a flow-matching feed-forward dimension of 4096. The architecture incorporates 32 joint diffusion transformer layers combined with multimodal RoPE. For text encoding, we employ a 1024-dimensional TTS text encoder, while the caption encoder utilizes Qwen2.5-7B~\citep{yang2025qwen2}. The Audio-VAE processes input waveforms at 44.1 kHz and generates embeddings at 43 Hz, achieving a 1024$\times$ downsample ratio relative to the input sampling rate. The Video-VAE handles input videos with varying resolutions and frame rates, producing embeddings at 3 Hz with 16× compression applied to both height and width dimensions. We train the model using the Adam optimizer with an initial learning rate of 1e-4.

% \Method comprises 26B parameters with a 4096-dimensional flow-matching layer. It features 32 joint diffusion transformer layers with RoPE. For text encoding, we use a 1024-dimensional TTS encoder, and the caption encoder is based on Qwen2.5-7B~\citep{yang2025qwen2}. The Audio-VAE processes 44.1 kHz waveforms, generating embeddings at 43 Hz with a 1024× downsample. The Video-VAE handles videos of varying resolutions and frame rates, producing embeddings at 3 Hz with 16× spatial compression. 
% The model is trained with the Adam optimizer at an initial learning rate of 1e-4.

\textbf{Baseline Methods.}
We select two canonical types of methods, including (1) Cascaded generation: these methods typically employ sequential T2V+V2A or T2A+A2V. Here, we employ AudioLDM2~\citep{liu2024audioldm} to perform the T2A task, while OpenSora~\citep{opensora} for T2V. (2) Joint generation. Recent works like JavisDiT~\citep{liu2025javisdit}, UniVerse-1~\citep{wang2025universe} and Ovi~\citep{low2025ovi} employ a dual-tower architecture with dedicated interaction layers between them.

\textbf{Evaluation Metrics.}
% To evaluate generation capabilities, we test T2AV tasks on video, audio, and consistency. Using Verse-Bench, we report the Motion Score (MS) for video dynamics, Aesthetic Score (AS) for visual fidelity, and ID Consistency (ID) for identity preservation. Audio quality is measured by Fréchet Distance (FD) and KL Divergence on mel-spectrograms, with semantic alignment evaluated via CLAP score. Audio-video synchronization is assessed with AV-A distance (Synchformer) and lip sync with SyncNet Confidence (SNC). Global alignment is measured by ImageBind (IB).
To evaluate the model’s generation capabilities, we test T2AV tasks on video, audio, and audio-video consistency. Following Universe-1, we use Verse-Bench for T2AV tasks. For video quality, we report the Motion Score (MS) based on RAFT~\citep{teed2020raft} optical flow for dynamic realism and the Aesthetic Score (AS), a composite metric from MANIQA~\citep{yang2022maniqa}, aesthetic-predictor-v2-5~\citep{aesthetic-predictor-v2-5}, and Musiq~\citep{ke2021musiq} for visual fidelity. Identity preservation is measured by ID Consistency (ID), calculated via DINOV3~\citep{simeoni2025dinov3} feature similarity. For audio quality, we use Fréchet Distance (FD) and KL Divergence on mel-spectrograms from PANNs~\citep{kong2020panns}. Semantic alignment is evaluated using the CLAP score~\citep{wu2023large}. For synchronization, we report AV-A (Audio-Video Alignment) distance from Synchformer~\citep{iashin2024synchformer} and SyncNet Confidence (SNC) score~\citep{chung2016out} for lip sync. Global cross-modal alignment is measured by ImageBind (IB).

\begin{table*}[!t]
\setlength\tabcolsep{5pt}
\centering
\renewcommand{\arraystretch}{1}
\caption{Main comparisons of audio-visual joint generation.}
\vspace{-5pt}
\label{tab:main-table}
\resizebox{1.0\linewidth}{!}{
\begin{tabular}{@{}lccccccccccc@{}}
\toprule
\multirow{2}{*}{\textbf{Method}} & \multirow{2}{*}{\textbf{Framework}} & \multicolumn{3}{c}{\textbf{Video}} & \multicolumn{3}{c}{\textbf{Audio}} & \textbf{TTS} & \multicolumn{3}{c}{\textbf{AV Consistency}} \\ \cmidrule(l){3-5} \cmidrule(l){6-8} \cmidrule(l){9-9} \cmidrule(l){10-12} 
 &  & \textbf{MS $\uparrow$} & \textbf{AS $\uparrow$} & \textbf{ID $\uparrow$} & \textbf{FD $\downarrow$} & \textbf{KL $\downarrow$} & \textbf{CLAP $\uparrow$} & \textbf{WER $\downarrow$} & \textbf{AV-A $\downarrow$} & \textbf{SNC $\uparrow$} & \textbf{IB-Score $\uparrow$} \\ \midrule
AudioLDM+TemoTkn & T2A+A2V & 0.05 & 0.12 & 0.07 & 2.05 & 2.53 & 0.229 & - & 1.15 & 2.68 & 0.137 \\ \midrule
OpenSora+FoleyGen & \multirow{2}{*}{T2V+V2A} & 0.42 & 0.44 & 0.56 & 3.69 & 3.08 & 0.212 & - & 0.92 & 2.76 & 0.159 \\
OpenSora+See\&Hear &  & 0.42 & 0.44 & 0.56 & 2.26 & 2.97 & 0.206 & 0.337 & 0.86 & 2.85 & 0.164 \\ \midrule
JavisDiT & \multirow{5}{*}{T2AV} & 0.18 & 0.36 & 0.22 & 1.95 & 5.17 & 0.228 & 0.256 & 0.92 & 3.94 & 0.231 \\
SVG &  & 0.40 & 0.41 & 0.25 & 1.55 & 3.62 & 0.080 & - & 0.72 & 4.07 & 0.206 \\
Universe-1 &  & 0.20 & 0.47 & 0.25 & 1.55 & 1.25 & 0.160 & 0.180 & 0.98 & 3.92 & 0.198 \\
Ovi &  & \textbf{0.58} & 0.48 & 0.46 & 1.50 & 1.19 & 0.224 & 0.035 & 0.82 & 4.28 & 0.214 \\ \midrule
\RC{30}\Method (Ours) & Unified T2AV & 0.48 & \textbf{0.51} & \textbf{0.59} & \textbf{1.36} & \textbf{1.06} & \textbf{0.232} & \textbf{0.028} & \textbf{0.65} & \textbf{6.79} & \textbf{0.316} \\ \bottomrule
\vspace{-3pt}
\end{tabular}
}
\end{table*}

\subsection{Comparison with Existing Methods}
\label{sec:compare_sota}

\Method \textbf{demonstrates robust audio–video joint generation.}
In Table~\ref{tab:main-table}, \Method achieves state-of-the-art performance, surpassing the two prior methods by a large margin. Cascaded approaches perform poorly due to error accumulation and strong dependence on upstream generation quality, while existing joint models exhibit only moderate audio–video consistency and noticeable unimodal degradation. In contrast, \Method attains substantially better A/V consistency and synchronization, which we attribute to the unified single-tower architecture and omni-full attention mechanism.

\Method \textbf{effectively guarantees unimodal performance.}
We then assess unimodal performance. As shown in Table~\ref{tab:main-table}, \Method delivers high audio and video quality in joint generation—surpassing cascaded and joint baselines by 34\% and 18\%, respectively—while multitask training yields more generalizable representations. Moreover, \Method outperforms specialized T2A and T2V models on their respective tasks, indicating that leveraging complementary audio–visual knowledge strengthens unimodal representations and further improves single-modality generation quality.

\Method \textbf{consistently maintains a performance advantage across multiple tasks.}
Table~\ref{tab:main-table} provides a comprehensive evaluation of \Method across a broad suite of tasks—including TI2AV, TI2V, T2V, and T2A—with comparisons to task-specialized baselines. \Method delivers consistently strong results, matching or surpassing specialized state-of-the-art models.

\begin{figure*}[!t]
  \centering
    \includegraphics[width=\linewidth]{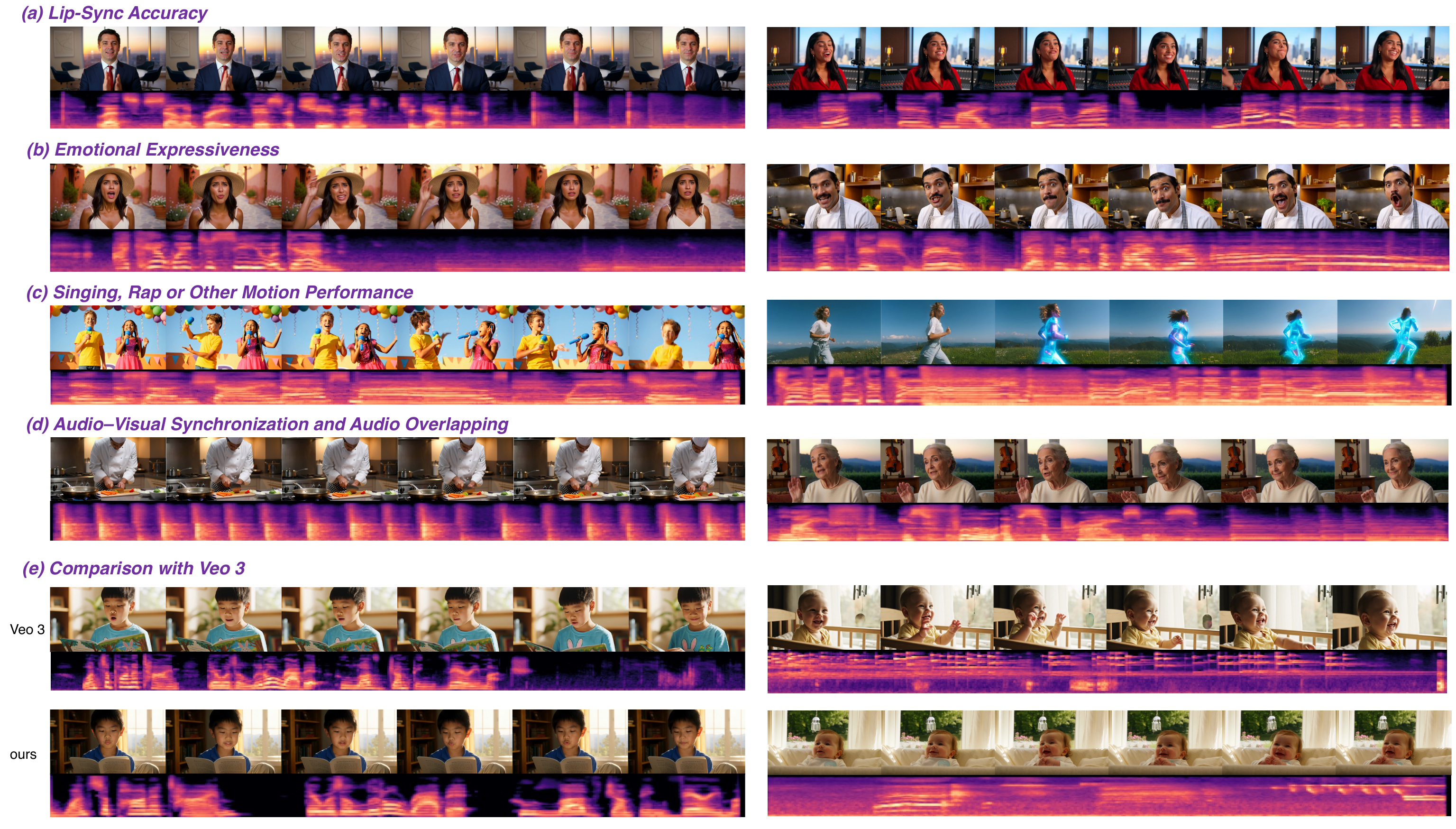}
    \vspace{-7pt}
   \caption{Qualitative evaluation of audio-video joint generation across various aspects.}
   \label{fig:quality}
\end{figure*}

\subsection{Qualitative Results}
\label{sec:visualize}

\textbf{Lip-Sync Accuracy.}
Fig.~\ref{fig:quality} (a–b) highlights lip-sync performance. \Method achieves phoneme-level alignment between mouth movements and audio—covering mouth openings, lip–teeth shapes, and tongue positions, while Universe-1 and Ovi suffer from misalignment, delay, and clear audio–visual mismatch.

\textbf{Emotional Expressiveness.}
From Fig.~\ref{fig:quality} (c), \Method generates characters with expressive emotions: facial cues (eyes, mouth curvature, muscle tension) are highly consistent with the audio’s affective tone (joy, sadness, excitement, lethargy), reflecting natural audio–visual fusion of prosody and dynamics. In contrast, Universe-1 and Ovi often produce distorted or emotional expressions.

\begin{table*}[!t]
    \setlength\tabcolsep{12pt}
    \centering
    \small
    \renewcommand{\arraystretch}{1}
    \caption{Comparison of different methods. The Dual Tower uses standard cross-attention, while the Single Tower utilizes our proposed Omni-Full Attention.}
    \vspace{-5pt}
    \label{tab:ablation_arch}
    \resizebox{.95\linewidth}{!}{
    \begin{tabular}{@{\hspace{6pt}}lccccccc@{\hspace{6pt}}}
        \toprule
        \textbf{Method} & \multicolumn{1}{c}{\textbf{Video}} & \multicolumn{2}{c}{\textbf{Audio}} & \multicolumn{1}{c}{\textbf{TTS}} & \multicolumn{3}{c}{\textbf{Audio-Video Consistency}} \\ 
        \cmidrule(l){2-2} \cmidrule(l){3-4} \cmidrule(l){5-5} \cmidrule(l){6-8}
        & \textbf{ID $\uparrow$} & \textbf{MOS $\uparrow$} & \textbf{CLAP $\uparrow$} & \textbf{WER $\downarrow$} & \textbf{DeSync $\downarrow$} & \textbf{Sync-conf $\uparrow$} & \textbf{IB $\uparrow$} \\
        \midrule
        Dual Tower & 0.62 & 62.02 & 0.139 & 0.675 & 1.163 & 3.762 & 0.126 \\
        Single Tower & 0.80 & 93.11 & 0.232 & 0.028 & 0.650 & 6.787 & 0.316 \\
        \bottomrule
    \end{tabular}
    }
\end{table*}

\textbf{Singing and Rap Performance.}
As shown in Fig.~\ref{fig:quality}, \Method yields performances where pitch, rhythm, and breath control are tightly aligned across audio and visuals—vibrato, melisma, and dynamic changes naturally match breathing patterns and facial expressions, consistent with human expectations of singing. In contrast, Universe-1 and Ovi show clear lip-sync failures, resulting in pronounced incongruity and reduced realism, especially for rap.

\textbf{Audio–Visual Synchronization and Audio Overlapping.}
As shown in Fig.~\ref{fig:quality}, \Method jointly generates background music and sound effects that are emotionally consistent with the video, with synchronized timing, realistic acoustics, and high fusion, thereby enhancing immersion. In contrast, baseline methods struggle to produce overlapping sounds and exhibit poor audio–visual consistency.

\textbf{Image to Audio-Video.}
Fig.~\ref{fig:quality} presents TI2AV and TI2V results. Our model preserves high identity consistency with the input image while generating plausible camera motion and dynamics, whereas baselines exhibit identity drift, large visual discrepancies, and mechanical movements.

\subsection{Ablation and Analysis}
\label{sec:ablation}

\textbf{Architectural Effectiveness.}
% To investigate the performance gap between single- and dual-tower architectures for audio–video generation, we feed audio and video features into separate mm-DiT branches, with the audio tower randomly initialized due to the lack of a matching DiT backbone. Each block uses full attention, and cross-attention is applied at every layer for feature alignment. We further conduct an ablation with a pretrained audio tower following the same training protocol. Both quantitative metrics and feature visualizations are evaluated. As shown in Table N, the single-tower model significantly outperforms the dual-tower variant in audio quality, video quality, and audio–video consistency; visualizations confirm that the single tower facilitates stronger cross-modal alignment, leading to substantial performance gains. Although the pretrained audio tower converges faster, its final performance is suboptimal, likely because excessive prior knowledge induces a distribution mismatch with video features, hindering effective alignment.
To compare single- and dual-tower architectures for audio–video generation, we feed audio and video features into separate mm-DiT branches, with a randomly initialized audio tower due to the lack of a matching DiT backbone. Each block uses full attention and cross-attention for feature alignment. We also conduct an ablation with a pretrained audio tower. As shown in Table ~\ref{tab:ablation_arch}, the single-tower model outperforms the dual-tower variant in audio quality, video quality, and audio–video consistency, with visualizations confirming better cross-modal alignment. Although the pretrained audio tower converges faster, its performance is suboptimal due to a distribution mismatch with video features, hindering alignment.

\begin{table*}[!t]
    \setlength\tabcolsep{13pt}
    \centering
    \renewcommand{\arraystretch}{1}
   \caption{Ablation of multi-task masking, with arrows indicating optimization directions.}
   % \vspace{-5pt}
  \label{tab:ablation_masking}
  \resizebox{.9\linewidth}{!}{
  \begin{tabular}{@{\hspace{6pt}}lccccccc@{\hspace{6pt}}}
    \toprule
    \textbf{Method} & \multicolumn{1}{c}{\textbf{Video}} & \multicolumn{2}{c}{\textbf{Audio}} & \multicolumn{1}{c}{\textbf{TTS}} & \multicolumn{3}{c}{\textbf{Audio-Video Consistency}} \\ \cmidrule(l){2-2} \cmidrule(l){3-4} \cmidrule(l){5-5} \cmidrule(l){6-8}
    & \textbf{ID $\uparrow$} & \textbf{MOS $\uparrow$} & \textbf{CLAP $\uparrow$} & \textbf{WER $\downarrow$} & \textbf{DeSync $\downarrow$} & \textbf{Sync-conf $\uparrow$} & \textbf{IB $\uparrow$} \\
    \midrule
    T2V & 0.71 & - & - & - & - & - & - \\
    T2V+T2AV & 0.76 & 88.181 & 0.188 & 0.044 & 0.895 & 5.024 & 0.201 \\
    All Tasks(Ours) & 0.80 & 93.106 & 0.232 & 0.028 & 0.650 & 6.787 & 0.316 \\
    \bottomrule
  \end{tabular}
  }
\end{table*}

\textbf{Advantages of Multi-Task Masking.}
As shown in Table~\ref{tab:ablation_masking}, for T2AV joint generation, our multi-task model significantly outperforms a counterpart trained solely on T2AV. This approach captures cross-modal audio–video correlations and complementary cues, outperforming video-only models on T2V and I2V. The unified multi-task training also produces robust, generalizable representations that scale well with data and compute, and the model generalizes effectively to I2AV and I2V, demonstrating high transferability.

% \begin{table}[!t]
%     \setlength\tabcolsep{1pt}
%     \centering
%     \renewcommand{\arraystretch}{0.9}
%   \caption{\textbf{Ablations of different training stages.} Metrics include video, audio, TTS, and audio-video consistency, with arrows indicating optimization directions.}
%   \vspace{-5pt}
%   \label{tab:training_stages}
%   \resizebox{\linewidth}{!}{
%   \begin{tabular}{@{\hspace{6pt}}lccccccc@{\hspace{6pt}}}
%     \toprule
%     \textbf{method} & \multicolumn{1}{c}{\textbf{Video}} & \multicolumn{2}{c}{\textbf{Audio}} & \multicolumn{1}{c}{\textbf{TTS}} & \multicolumn{3}{c}{\textbf{Audio-Video Consistency}} \\ \cmidrule(l){2-2} \cmidrule(l){3-4} \cmidrule(l){5-5} \cmidrule(l){6-8}
%     & \textbf{ID$\uparrow$} & \textbf{MOS$\uparrow$} & \textbf{CLAP $\uparrow$} & \textbf{WER$\downarrow$} & \textbf{DeSync$\downarrow$} & \textbf{Sync-Conf$\uparrow$} & \textbf{IB$\uparrow$} \\
%     \midrule
%     (a) pretrain & 0.52 & 89.602 & 0.165 & 0.043 & 1.030 & 6.431 & 0.163 \\
%     (b) post-train-special & 0.68 & 90.72 & 0.210 & 0.035 & 0.995 & 6.472 & 0.263 \\
%     (c) post-train-quality & 0.80 & 93.106 & 0.232 & 0.028 & 0.600 & 6.787 & 0.316 \\
%     \bottomrule
%   \end{tabular}
%   }
%   \vspace{-3pt}
% \end{table}

\begin{figure}[!t]
    \centering
    \includegraphics[width=0.9\linewidth]{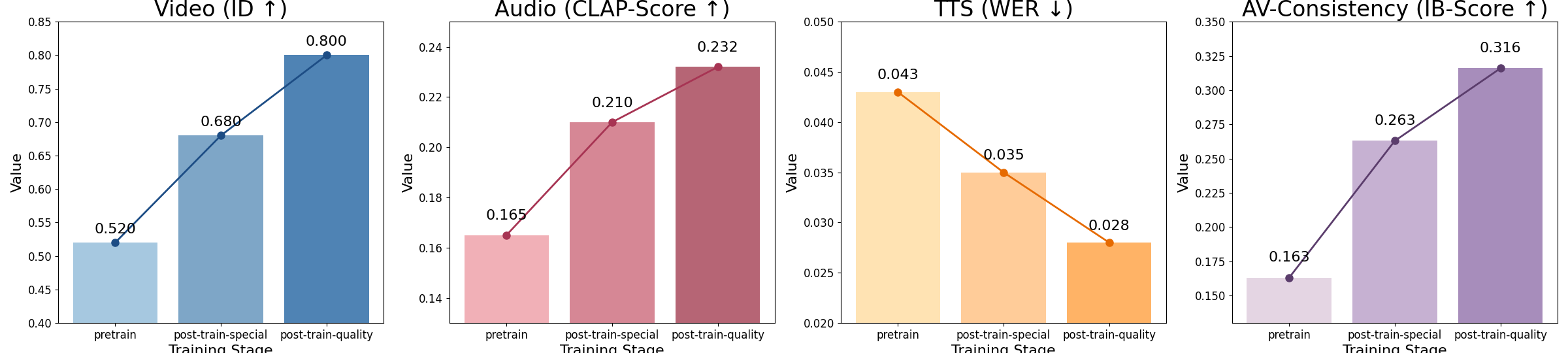}
    \caption{\textbf{Ablations of different training stages.} Metrics include video, audio, TTS, and audio-video consistency, with arrows indicating optimization directions.}
    \label{fig:abla_stage}
\end{figure}

\textbf{Gains from Progressive Training Strategy.}
As in Figure~\ref{fig:abla_stage}, the progressive training first equips the model with basic audio–video generation capabilities (a). In the post-training phase, the full model. The SP stage further boosts previously weak skills (\eg, IB score by 0.1), and post-training on high-quality data brings an additional overall gain over (b). Removing the entire progressive schedule instead causes a significant drop, confirming the effectiveness of our multi-stage training strategy.

% \textbf{Benefits of MixD-RoPE.}
% We train at native FPS without resampling audio or video, allowing the model to handle arbitrary sampling-rate combinations. As shown in Table N, our RoPE variant outperforms alternative designs, likely because it accommodates dynamic audio–video FPS without coupling or resampling, thereby modeling native frame rates and enabling dynamic temporal scaling for alignment.

%% file: sec/6_conclusion.tex
\section{Conclusion}
\label{sec:conclusion}
% In this paper, we identify persistent failure modes in audio–video generation, \eg, audio-video asynchrony, lip–speech mismatch, and degradation of unimodal generation, arising from suboptimal architectures, insufficiently aligned data, and single-task training paradigms. To solve these issues, we introduce \Method, a unified multi-task framework with a single-tower backbone and Omni-Full Attention for full-dimensional cross-modal interaction, a progressive multi-task regime with random modality masking and performance-adaptive curricula, and an automated annotation pipeline. This pipeline yields a large-scale, high-quality, densely annotated audio–video dataset that is crucial for cross-modal alignment and scalable training. Our analyses examine the gap between single- and dual-tower designs, the impact of audio-tower pretraining, the fusion behavior of Omni-Full Attention, and the gains from progressive multi-task learning. \Method surpasses prior state-of-the-art on both unimodal and joint audio-video benchmarks, becoming the first open-source model comparable to Veo 3.
We identify key failure modes in audio–video generation, such as asynchrony, lip–speech mismatch, and unimodal degradation, caused by suboptimal architectures, misaligned data, and single-task training. To address these, we propose \Method, a unified multi-task framework with a single-tower backbone and Omni-Full Attention for cross-modal interaction, alongside a progressive training strategy and an automated annotation pipeline. This pipeline produces a high-quality, annotated audio–video dataset for scalable training. \Method outperforms prior state-of-the-art on unimodal and joint audio-video benchmarks, becoming the superior model comparable to Veo 3.
We hope this work could provide a clear direction and catalyze deeper research in audio–video generation.

\section*{Acknowledgement}
We acknowledge the contributions from (sorted by first name): Boyuan Jiang, Chen Zhang, Feng Deng, Haorui Zheng, Jiachen Zheng, Jiahao Wang, Jiahui Zhao, Jingbin He, Jingke Li, Jingru Zhao, Junjie Yan, Kang Yin, Le Wang, Liang Hou, Lingyu Zou, Ming Wen, Nan Li, Peihan Li, Pengfei Cai, Pengfei Wan, Qianyue Hu, Shiyao Wang, Teng Ma, Xiaopeng Wang, Xiaoyu Shi, Xin Tao, Xu Li, Yan Zhou, Youjun Chen, Yu Zhao, Yuan Gao, Yuejiao Wang, Yun Li, Yushen Chen, Yuzhe Liang, Zewen Song, Zhongliang Liu, Zihan Li, Zihao Ji, Ziyang Yuan, and Ziyu Zhang.